\documentclass{article}

\usepackage{arxiv}
\usepackage[utf8]{inputenc}
\usepackage[T1]{fontenc}
\usepackage{microtype}
\usepackage{amsmath,amssymb,amsthm,mathtools}
\usepackage{booktabs}
\usepackage{graphicx}
\usepackage{xcolor}
\usepackage{hyperref}
\usepackage{natbib}
\usepackage{enumitem}
\usepackage{subfig}

\hypersetup{
  hidelinks,
  pdftitle={Decentralized Causal Discovery Using Judo Calculus},
  pdfauthor={Sridhar Mahadevan},
  pdfsubject={Multi-environment causal discovery},
  pdfkeywords={causal discovery, multi-environment learning, stability selection, sheaves, do-calculus}
}

\newtheorem{proposition}{Proposition}

\newcommand{\doop}{\operatorname{do}}
\newcommand{\Pa}{\operatorname{Pa}}
\newcommand{\indep}{\perp\!\!\!\perp}
\newcommand{\E}{\mathcal E}
\newcommand{\C}{\mathcal C}
\newcommand{\J}{J}

\newcommand{\A}{\mathcal A}
\newcommand{\1}{\mathbf 1}

\title{Decentralized Causal Discovery Using Judo Calculus}

\author{
  Sridhar Mahadevan\\
  Adobe Research and University of Massachusetts Amherst\\
  \texttt{smahadev@adobe.com, mahadeva@umass.edu}
}

\begin{document}
\maketitle

\begin{abstract}
We study decentralized causal discovery from data partitioned across regimes,
environments, laboratories, or interventions.  A base learner is applied locally,
and its graph, conditional-independence, or mechanism estimates are combined by a
pre-specified stability rule.  We call this cover-indexed procedure
\emph{\(J\)-stable discovery}, or informally ``judo calculus.''  The terminology is
motivated by the companion theory of \(j\)-do-calculus, in which ordinary
do-calculus identities established in compatible local causal models descend
across a cover.  The algorithms in this paper are statistical approximations to
that local-to-global pattern; frequency aggregation by itself is not a proof of a
do-calculus premise.

We instantiate the procedure with score-based, constraint-based, and
gradient-based learners, including GES, \(\psi\)-FCI, and DCDI.  We distinguish
ordinary support aggregation from an intervention-aware version that excludes
regimes in which the child mechanism was directly replaced.  Experiments on
synthetic linear structural causal models, the Sachs protein-signaling data,
LINCS perturbation signatures, and OECD PISA variables illustrate the
precision--recall and computational tradeoffs.  The results support a modest
conclusion: decentralized stability filtering can remove brittle edges and can
be parallelized effectively, but its causal interpretation depends on the
choice of regimes, intervention masks, local learner, and compatibility
assumptions.
\end{abstract}


\section{Introduction}
\label{sec:intro}

Causal discovery is often performed on a single pooled table, even when the data
come from distinct experimental conditions, countries, cell lines, doses, time
points, or institutions.  Pooling can increase sample size, but it can also mix
different mechanisms and conceal which conclusions persist across regimes.
Conversely, fitting each regime independently exposes heterogeneity but leaves
the problem of combining many noisy local graphs.

This paper investigates a simple decentralized alternative:
\begin{enumerate}[leftmargin=*,itemsep=2pt]
  \item choose an admissible family of regimes for the question of interest;
  \item run a standard causal-discovery method locally in each regime;
  \item retain edges, independences, or mechanisms that receive sufficient
        compatible support across the family; and
  \item report the local evidence and the aggregation threshold together with
        the resulting graph.
\end{enumerate}
We call the admissible family a \emph{cover} and the resulting decision
\emph{\(J\)-stable}.  Operationally, this is a multi-environment stability
selection method.  It is compatible with the sheaf-theoretic viewpoint that
motivates \(j\)-do-calculus, but the empirical aggregation rule should not be
confused with literal sheafification or with a universal Kan-extension
construction.

The distinction matters.  The companion \(j\)-do-calculus paper
\citep{sm:judo_calc} proves a local-to-global result under explicit hypotheses:
site-indexed structural causal models and graph surgeries must be compatible
under restriction; the ordinary Pearl premise must hold in every relevant
chart; conditionals must be defined; and the resulting local identities must
agree on overlaps.  In contrast, an edge-frequency threshold is evidence about
repeatability.  It does not, without further assumptions, certify
d-separation, identify an adjustment set, or orient a causal edge.

Our contributions are therefore empirical and algorithmic:
\begin{itemize}[leftmargin=*,itemsep=2pt]
  \item a method-agnostic framework for decentralized graph and
        conditional-independence aggregation;
  \item an intervention-aware support statistic that does not penalize a true
        incoming edge merely because an experiment replaced its child
        mechanism;
  \item score-, constraint-, and gradient-based realizations using GES,
        \(\psi\)-FCI, and DCDI;
  \item a finite-sample concentration statement for support thresholding under
        explicit repeated-selection assumptions; and
  \item an empirical study showing both successes and limitations of the
        approach.
\end{itemize}

The present version deliberately leaves the categorical theory to
\citet{sm:judo_calc}.  No conditioning--intervention duality via left and right
Kan extensions is assumed or required here.

\section{Cover-Indexed Stability}
\label{sec:framework}

\subsection{Regimes, covers, and local outputs}

Let \(V=\{X_1,\ldots,X_d\}\) be a common variable set and let
\(\E=\{e_1,\ldots,e_m\}\) index environments.  An environment may be
observational, interventional, or simply a population or measurement regime.
For a query or child variable \(X_j\), a \emph{cover}
\(\J_j\subseteq\E\) is the pre-specified family regarded as comparable and
admissible for that decision.  Different children may use different covers.
For example, if \(X_j\) is directly intervened upon in environment \(e\), then
\(e\) should usually be excluded when assessing invariance of the mechanism for
\(X_j\).

Let a base learner \(\A\) return a local graph summary
\[
  A^{(e)}\in\{0,1\}^{d\times d},\qquad
  A^{(e)}_{ij}=1
  \ \Longleftrightarrow\
  \A\text{ selects }X_i\to X_j\text{ in environment }e.
\]
For a CPDAG or PAG, we separately retain skeleton and orientation marks rather
than silently converting every adjacency into a directed edge.

The regimes can be organized as objects of a site \((\C,\J)\), with restriction
maps representing refinements and \(\J\)-covers representing admissible local
families.  This organization is useful when overlaps are genuine and local
objects have defined restriction maps.  The statistical algorithms below need
only the finite cover and explicit compatibility checks; calling an arbitrary
collection of data partitions a cover does not by itself make the outputs a
sheaf.

\subsection{Support aggregation}

For noninterventional regimes, the unweighted directed support is
\[
  F_{ij}=\frac{1}{|\J_j|}\sum_{e\in\J_j}A^{(e)}_{ij}.
  \label{eq:support}
\]
At a threshold \(\pi\in[0,1]\), we retain \(i\to j\) when
\(F_{ij}\ge\pi\).  The choices \(\pi=1\), \(\pi=1/m\), and
\(\pi=1-k/m\) correspond to intersection, union, and all-but-\(k\)
aggregation.  Intersection is conservative; union is diagnostic; an
intermediate threshold trades false positives against false negatives.

Directions are less stable than adjacencies.  We therefore first define
\[
  F^{\mathrm{skel}}_{\{i,j\}}
  =\frac{1}{|\J_{ij}|}\sum_{e\in\J_{ij}}
    \1\!\left[A^{(e)}_{ij}=1\ \text{or}\ A^{(e)}_{ji}=1\right],
\]
and orient only when the base learner supplies orientation evidence and the net
margin \(F_{ij}-F_{ji}\) exceeds a separately reported \(\delta>0\).
Majority direction is a heuristic, not a substitute for valid orientation
rules.

\subsection{Intervention-aware support}
\label{sec:masked}

Perfect intervention on \(X_j\) removes incoming arrows into \(X_j\) from the
mutilated data-generating graph in that regime.  Consequently, literal
intersection across all interventional regimes can delete genuine edges in the
underlying pre-intervention DAG.

Let \(m_{ej}=1\) when environment \(e\) leaves the mechanism of \(X_j\)
unchanged and \(m_{ej}=0\) when it directly intervenes on \(X_j\).  With
pre-specified nonnegative regime weights \(w_e\), define
\begin{equation}
  F^{\mathrm{mask}}_{ij}
  =
  \frac{\sum_{e\in\J_j}w_e m_{ej}A^{(e)}_{ij}}
       {\sum_{e\in\J_j}w_e m_{ej}},
  \qquad
  \sum_{e\in\J_j}w_em_{ej}>0.
  \label{eq:masked}
\end{equation}
Unknown targets can be handled only by an additional target-estimation model or
by reporting sensitivity to plausible masks.  Treating the absence of an edge
under a direct intervention as ordinary negative evidence is generally
incorrect.

\subsection{What thresholding can guarantee}

The following elementary result states the statistical content of repeated
support.  It does not claim causal identifiability.

\begin{proposition}[Support separation]
\label{prop:concentration}
Fix an eligible edge \(i\to j\).  Suppose its local selection indicators over
\(m_j\) eligible environments are independent Bernoulli variables.  Assume a
true edge is selected in each environment with probability at least \(p_T\),
whereas a false edge is selected with probability at most \(p_F\), and choose
\(p_F<\pi<p_T\).  Then
\[
 \Pr(F^{\mathrm{mask}}_{ij}<\pi\mid i\to j\text{ true})
 \le e^{-2m_j(p_T-\pi)^2},
\]
and
\[
 \Pr(F^{\mathrm{mask}}_{ij}\ge\pi\mid i\to j\text{ false})
 \le e^{-2m_j(\pi-p_F)^2}.
\]
\end{proposition}

\begin{proof}
Apply Hoeffding's inequality to the mean of the eligible selection indicators.
\end{proof}

The assumptions are strong: regimes may be dependent, \(p_T\) can vary by
mechanism, and interventions can alter detectability.  Under dependence one
needs an appropriate concentration bound or a regime-level bootstrap.  In
particular, the proposition does not justify an all-but-\(k\) rule with
\(k=o(m)\) when every local learner has a fixed nonzero false-negative rate.

\subsection{Relation to \(j\)-do-calculus}

In the companion theory, a compatible family of causal models assigns a local
SCM, mutilated graph, and probability law to each chart.  If one of Pearl's
three do-calculus rules applies in every chart of a common cover and the
resulting conditional distributions agree on overlaps, separatedness of the
probability sheaf yields the corresponding identity at the covered stage
\citep{pearl-book,sm:judo_calc}.  This is the sound theoretical interface used
here.

Our algorithms can help search for stable candidate premises, but the logical
direction is one way:
\[
\text{compatible chartwise causal proof}
\Longrightarrow
\text{local-to-global \(j\)-do identity}.
\]
The converse
\[
\text{high empirical edge frequency}
\Longrightarrow
\text{do-calculus premise}
\]
does not hold without additional assumptions.

When regime-specific causal effects \(P^e(Y\in B\mid\doop(X=x))\) have been
identified chartwise, they should be reported as a family, a range, or a
pre-specified mixture.  For weights \(\lambda_e\ge0\) with
\(\sum_e\lambda_e=1\),
\[
 P^{\mathrm{mix}}(Y\in B\mid\doop(X=x))
 =
 \sum_{e\in\J}\lambda_e
 P^e(Y\in B\mid\doop(X=x))
\]
is again a probability measure.  Arbitrary monotone functions, Fisher
combination, and Stouffer combination are not probability aggregators and are
not used to define an interventional distribution.

\section{Decentralized Discovery Algorithms}
\label{sec:algorithms}

\subsection{Generic procedure}

The common wrapper around a base learner is:
\begin{enumerate}[leftmargin=*,itemsep=3pt]
  \item \textbf{Specify regimes and eligibility.}  Record the environment
        labels, admissible covers, known intervention targets, and masks
        \(m_{ej}\).
  \item \textbf{Fit locally.}  Run the same learner and preprocessing in each
        eligible regime, retaining weighted graphs, CPDAGs, PAGs, or CI
        statistics.
  \item \textbf{Align outputs.}  Put graphs on the common variable set and
        distinguish missing variables from absent edges.
  \item \textbf{Aggregate.}  Compute skeleton and directed support using
        Eq.~\eqref{eq:masked}; use an explicit \(\pi\) and, for direction, an
        explicit margin \(\delta\).
  \item \textbf{Validate.}  Select hyperparameters using training/validation
        environments without reference to the test graph.  Report pooled,
        per-regime, union, intersection, and intermediate-threshold results.
\end{enumerate}

The reducer costs \(O(md^2)\) and needs only graph summaries, which is useful
when raw records cannot leave an institution.  The local fits are parallel.
This provides a wall-clock advantage when workers are available, although total
compute can equal or exceed that of a pooled run.

\subsection{Score-based discovery}

GES uses a decomposable score such as Gaussian BIC to search over Markov
equivalence classes \citep{chickering:jmlr}.  We consider two empirical
modifications.

\paragraph{CGES.}
The structural variant augments the score with sparsity and triangle penalties:
\[
 \mathrm{Score}_{\mathrm{CGES}}(G)
 =
 \mathrm{BIC}(G)-\lambda_1 f_1(S(G))-\lambda_2 f_2(S(G)),
\]
where \(S(G)\) is the undirected skeleton, \(f_1\) counts edges, and \(f_2\)
counts triangles.  These are graph regularizers; their usefulness is empirical
and does not require a topos interpretation.

\paragraph{TCES.}
When environment labels are available, we additionally penalize variation of a
local mechanism:
\[
 \mathcal L_{\mathrm{inv}}(v;U)
 =
 \sum_{e<e':\,m_{ev}m_{e'v}=1}
 \left\|\widehat\beta^{(e)}_{v\mid U}
       -\widehat\beta^{(e')}_{v\mid U}\right\|_2 .
\]
The score is
\[
 \mathrm{Score}_{\mathrm{TCES}}(G)
 =
 \mathrm{BIC}(G)
 -\lambda_1 f_1(S(G))
 -\lambda_2 f_2(S(G))
 -\lambda_J\sum_v\mathcal L_{\mathrm{inv}}(v;\Pa_G(v)).
\]
For nonlinear mechanisms the coefficient difference can be replaced by a
cross-environment residual or conditional-distribution discrepancy.  This is an
invariance regularizer related to invariant causal prediction
\citep{icm}; it is not literal sheaf descent.

\subsection{Constraint-based discovery}

For each tested statement \(X_i\indep X_j\mid X_S\), let \(p_e(i,j,S)\)
be its local \(p\)-value.  There are two different goals:

\begin{itemize}[leftmargin=*,itemsep=2pt]
  \item To test a common null across eligible environments, use a calibrated
        combination method whose dependence assumptions are stated.
  \item To require nonrejection in every eligible environment, apply local
        multiplicity correction and use a conjunction rule.  Nonrejection is
        not proof that the CI is true; power or equivalence testing is needed
        for a positive invariance claim.
\end{itemize}

We insert the resulting oracle into FCI or \(\psi\)-FCI and separately aggregate
the returned PAG skeletons.  Fisher and Stouffer methods combine \(p\)-values,
not probabilities or truth values.  The experiments below primarily use
per-regime PAGs followed by support aggregation, which keeps the procedure
auditable.

\subsection{Gradient-based discovery}

For DCDI \citep{dcdi}, one option is post-hoc aggregation.  Train a model in
each regime, obtain weighted adjacency \(W^{(e)}\), apply a fixed threshold or
validation-selected sparsity, and compute Eq.~\eqref{eq:masked}.  A second
option is joint training with regime-specific heads:
\[
 \min_{\{W^{(e)}\}}
 \sum_e\mathcal L_{\mathrm{DCDI}}(W^{(e)};D^{(e)})
 +
 \lambda_J
 \sum_{i\ne j}\operatorname{Var}_{e:m_{ej}=1}
 \left[\sigma(W^{(e)}_{ij})\right].
\]
The mask is essential under direct interventions.  We report skeleton metrics
when directions are unstable and never use the ground-truth edge count to
select a model intended for non-oracle evaluation.

\section{Experimental Design}
\label{sec:design}

\subsection{Questions and metrics}

The experiments ask:
\begin{enumerate}[leftmargin=*,itemsep=2pt]
  \item Does support aggregation reduce brittle false-positive edges?
  \item How much recall is lost as \(\pi\) increases?
  \item What wall-clock parallelism is available, and what is the total work?
  \item How stable are learned structures on real multi-environment data without
        a known DAG?
\end{enumerate}

With synthetic ground truth we report precision, recall, \(F_1\), and structural
Hamming distance (SHD).  CPDAG and PAG outputs are evaluated as skeletons unless
an orientation is identifiable and represented by the output type.  On
real-world data without ground truth, we report support, Jaccard overlap,
validation likelihood, and qualitative structure rather than causal accuracy.

Hyperparameters must be chosen without the test graph.  A row-stratified
60/20/20 split evaluates new samples from observed regimes; holding out entire
regimes evaluates environment transfer.  These protocols are not
interchangeable and are named explicitly below.

\subsection{Synthetic models}

We generate linear-Gaussian SCMs on random DAGs.  A random topological order is
sampled, edges point forward in that order, nonzero weights have random signs
and magnitudes in \([0.5,2]\), and
\[
 X_j=\sum_{i\in\Pa(j)}W_{ij}X_i+\varepsilon_j,
 \qquad \varepsilon_j\sim\mathcal N(0,1).
\]
Interventional regimes replace the equation for a target \(t(e)\) by
\(X_{t(e)}=\mu_e+\varepsilon_{t(e)}\).  Unless otherwise stated, the DCDI study
uses one observational and several single-node intervention regimes, a total
sample budget of \(10{,}000\), and ten random graphs per condition.

The archived plots in this version were generated by the original unmasked
support implementation.  We retain them as empirical comparisons of that
estimator, but do not reinterpret unmasked intersection as a consistent
estimator of the common pre-intervention DAG.  Equation~\eqref{eq:masked} is the
corrected estimator for future target-aware runs.

\subsection{Real datasets}

\paragraph{Sachs.}
The Sachs data contain single-cell measurements of 11 signaling variables under
multiple perturbation conditions \citep{sachs}.  Conditions define regimes.
The published network is used only as a conventional reference, not as an
unquestionable ground truth.

\paragraph{LINCS.}
LINCS L1000 provides perturbation signatures across cell lines, doses, and time
points \citep{lincs1000}.  We use compact landmark-gene panels and treat
cell-line--time--dose combinations as regimes.  Because no gold-standard DAG is
available, the study is descriptive.

\paragraph{PISA.}
Countries define regimes in the OECD PISA ESCS extract.  The four-variable
example contains the ESCS composite and three of its components.  A domain guard
forbids arrows from the constructed composite to its components.  Recovering
component-to-composite relations is therefore a pipeline sanity check, not an
independent discovery of socioeconomic causation.

\section{Results}
\label{sec:results}

\subsection{Computational behavior}

Table~\ref{tab:runtime} reports the archived DCDI timing comparison at an equal
iteration budget.  Per-regime jobs can be distributed across workers, and the
graph reducer is negligible.  The numbers show wall-clock behavior for this
implementation; they do not establish a lower total computational complexity.

\begin{table}[t]
\centering
\caption{Archived CPU time in seconds per 10,000 iterations.}
\label{tab:runtime}
\begin{tabular}{rrrr}
\toprule
\(d\) & Vanilla & Decentralized & Ratio \\
\midrule
10 & 30.3 & 27.8 & 0.90 \\
20 & 51.8 & 42.3 & 0.82 \\
40 & 143.5 & 101.9 & 0.71 \\
\bottomrule
\end{tabular}
\end{table}

\begin{figure}[t]
\centering
\subfloat[Per-iteration time]{\includegraphics[width=.48\linewidth]{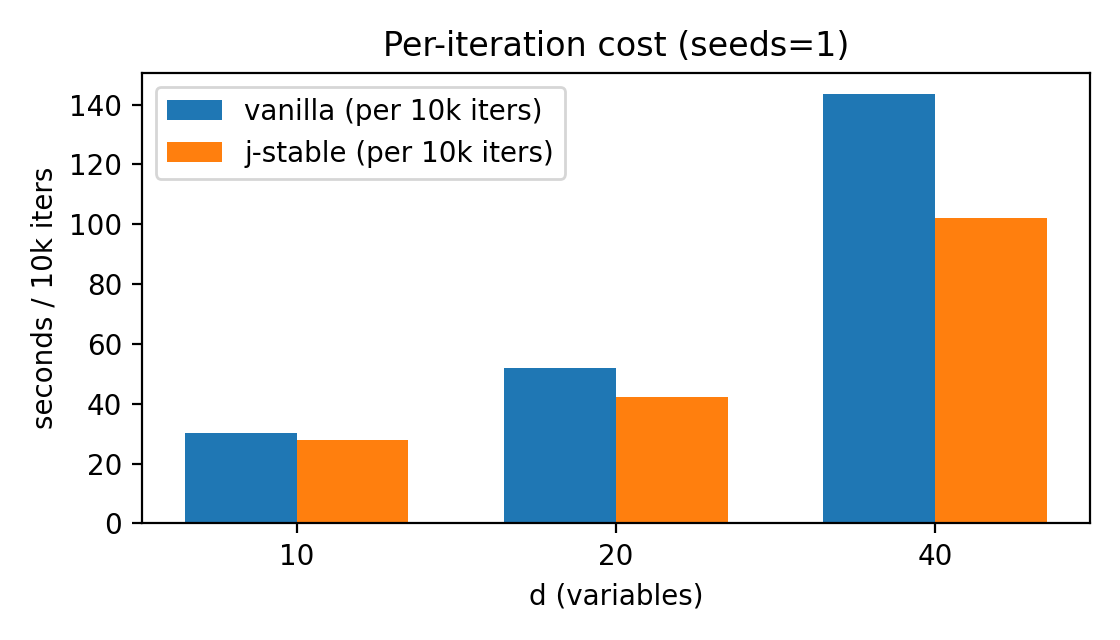}}
\hfill
\subfloat[Four-worker scaling]{\includegraphics[width=.48\linewidth]{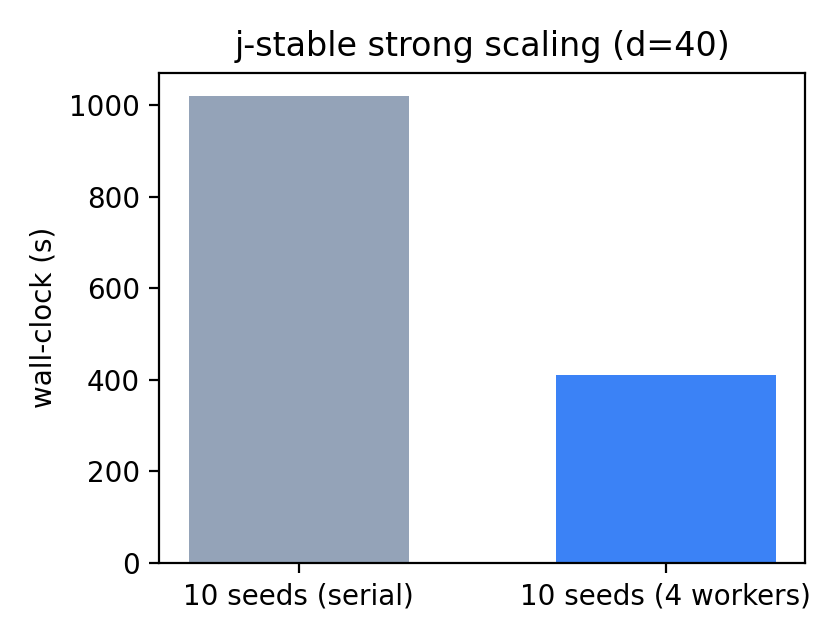}}
\caption{Archived timing measurements for vanilla and decentralized DCDI.
Parallel speedup depends on worker availability and regime balance.}
\label{fig:runtime}
\end{figure}

\subsection{Synthetic \(\psi\)-FCI and GES}

On one three-regime synthetic instance, intersection improves SHD but sacrifices
some recall.  Table~\ref{tab:fci-sweep} is a case study, not an average over
random graphs.  The nearly flat sweep indicates that aggregation, rather than
the tested \(\alpha\) values, drives the difference.

\begin{table}[t]
\centering
\caption{Skeleton \(F_1\) and SHD for pooled and decentralized \(\psi\)-FCI on
one synthetic three-regime instance.}
\label{tab:fci-sweep}
\begin{tabular}{rcccccc}
\toprule
& \multicolumn{2}{c}{Pooled}
& \multicolumn{2}{c}{Intersection}
& \multicolumn{2}{c}{All-but-1}\\
\(\alpha\) & \(F_1\) & SHD & \(F_1\) & SHD & \(F_1\) & SHD\\
\midrule
0.005 & 0.286 & 10 & 0.333 & 4 & 0.200 & 8\\
0.010 & 0.286 & 10 & 0.333 & 4 & 0.200 & 8\\
0.020 & 0.267 & 11 & 0.333 & 4 & 0.200 & 8\\
\bottomrule
\end{tabular}
\end{table}

\begin{figure}[t]
\centering
\subfloat[\(F_1\)]{\includegraphics[width=.48\linewidth]{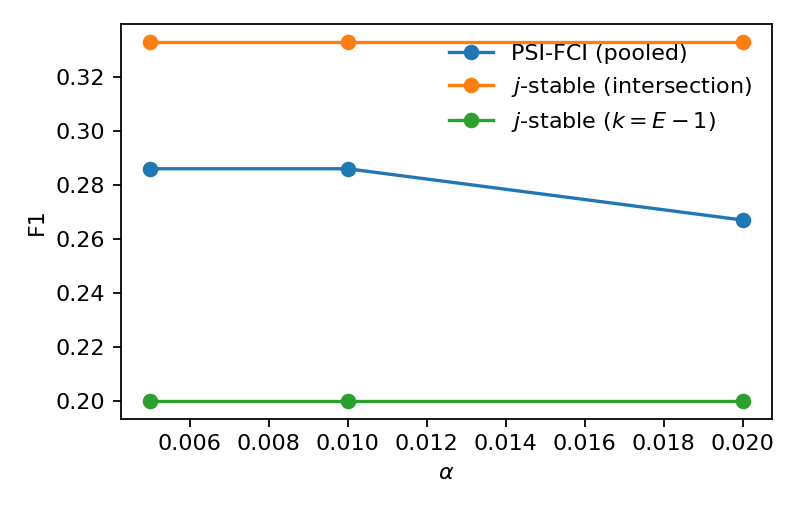}}
\hfill
\subfloat[SHD]{\includegraphics[width=.48\linewidth]{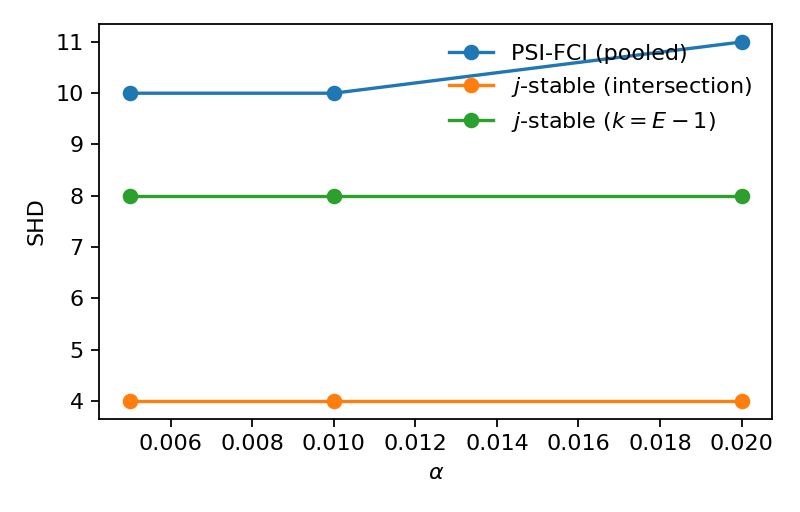}}
\caption{Sensitivity to the CI threshold on the synthetic case study.}
\label{fig:fci-sweep}
\end{figure}

The corresponding GES case study is shown in Table~\ref{tab:ges}.  Intersection
matches the six-edge true skeleton on this instance.  Because the same graph was
used to illustrate the procedure, this result should not be read as a general
consistency theorem.

\begin{table}[t]
\centering
\caption{GES skeleton comparison on one synthetic instance.}
\label{tab:ges}
\begin{tabular}{lrrrr}
\toprule
Method & Precision & Recall & \(F_1\) & SHD\\
\midrule
Pooled GES & 0.30 & 1.00 & 0.462 & 14\\
Intersection of local GES graphs & 1.00 & 1.00 & 1.000 & 0\\
\bottomrule
\end{tabular}
\end{table}

\subsection{Synthetic DCDI}

Table~\ref{tab:dcdi} summarizes the archived linear-perfect results over ten
random graphs per condition.  The decentralized estimator has lower median SHD
in these experiments, with the largest gap in the sparse settings.  Because the
archived implementation used unmasked graph frequencies, this is an empirical
performance result for that procedure rather than evidence that graph
intersection reconstructs a common DAG under arbitrary interventions.

\begin{table}[t]
\centering
\caption{Linear perfect-intervention experiments: median SHD \(\pm\) IQR over
ten random graphs.}
\label{tab:dcdi}
\resizebox{\linewidth}{!}{
\begin{tabular}{lrrrr}
\toprule
& \multicolumn{2}{c}{Directed SHD}
& \multicolumn{2}{c}{Skeleton SHD}\\
Condition & Vanilla & Decentralized & Vanilla & Decentralized\\
\midrule
\(d=10,e=1\) & \(22.5\pm6.75\) & \(6.0\pm1.75\)
             & \(22.5\pm6.75\) & \(6.0\pm2.50\)\\
\(d=20,e=1\) & \(100.5\pm7.00\) & \(13.5\pm5.25\)
             & \(100.5\pm7.00\) & \(12.5\pm4.50\)\\
\(d=20,e=4\) & \(106.0\pm8.25\) & \(76.0\pm7.00\)
             & \(102.0\pm2.75\) & \(72.5\pm8.25\)\\
\bottomrule
\end{tabular}}
\end{table}

\begin{figure}[t]
\centering
\subfloat[Directed SHD]{\includegraphics[width=.48\linewidth]{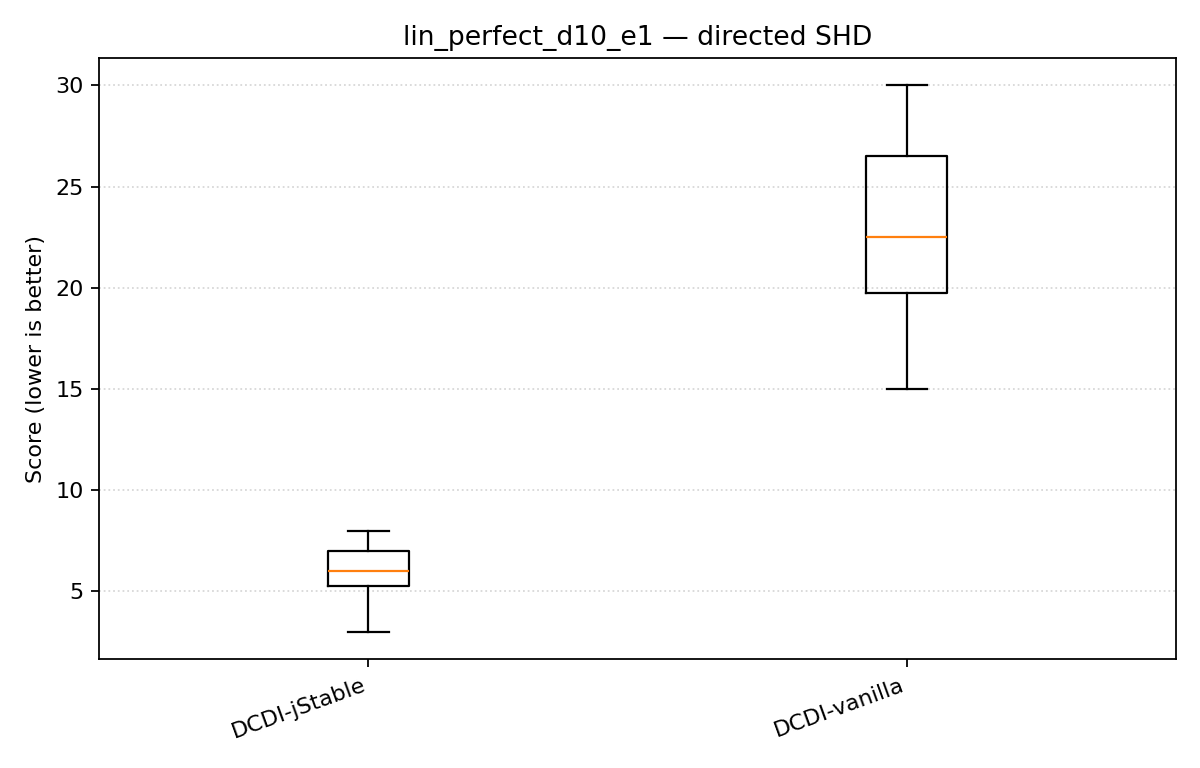}}
\hfill
\subfloat[Skeleton SHD]{\includegraphics[width=.48\linewidth]{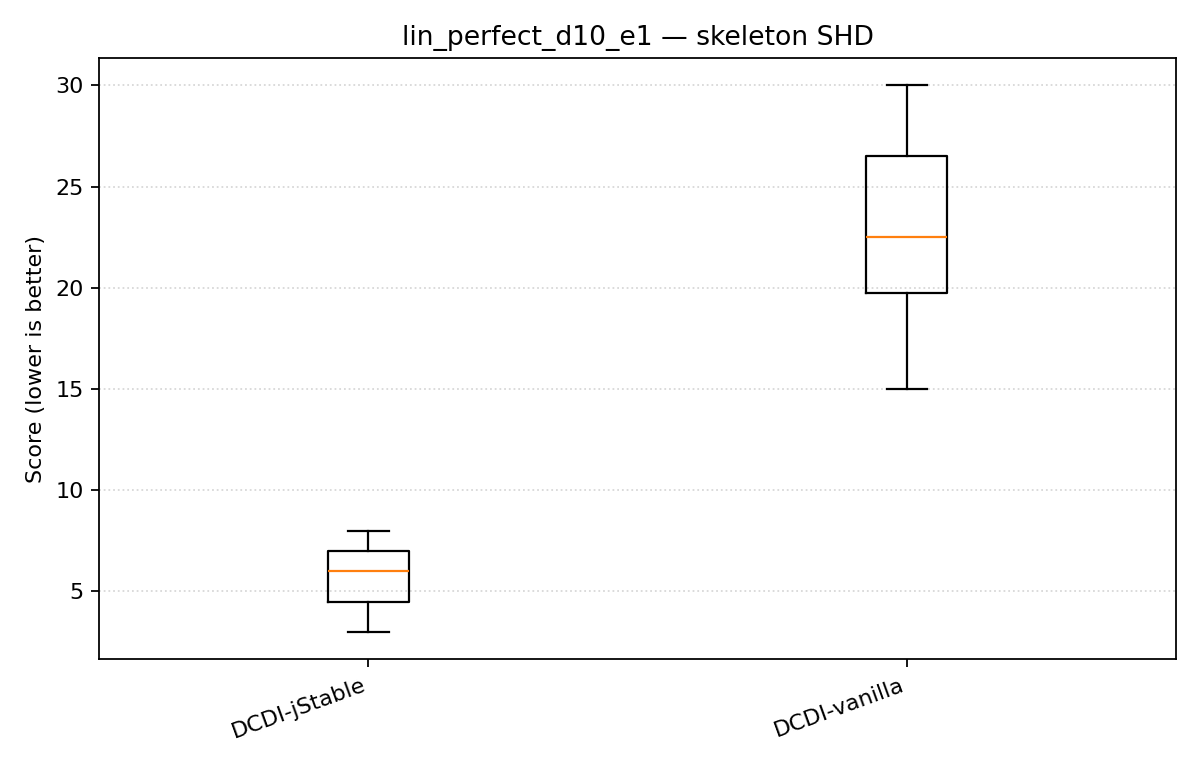}}
\caption{DCDI results for the \(d=10,e=1\) condition.  Boxplots summarize ten
random graphs.}
\label{fig:dcdi10}
\end{figure}

\begin{figure}[t]
\centering
\subfloat[\(d=20,e=1\), directed]{\includegraphics[width=.48\linewidth]{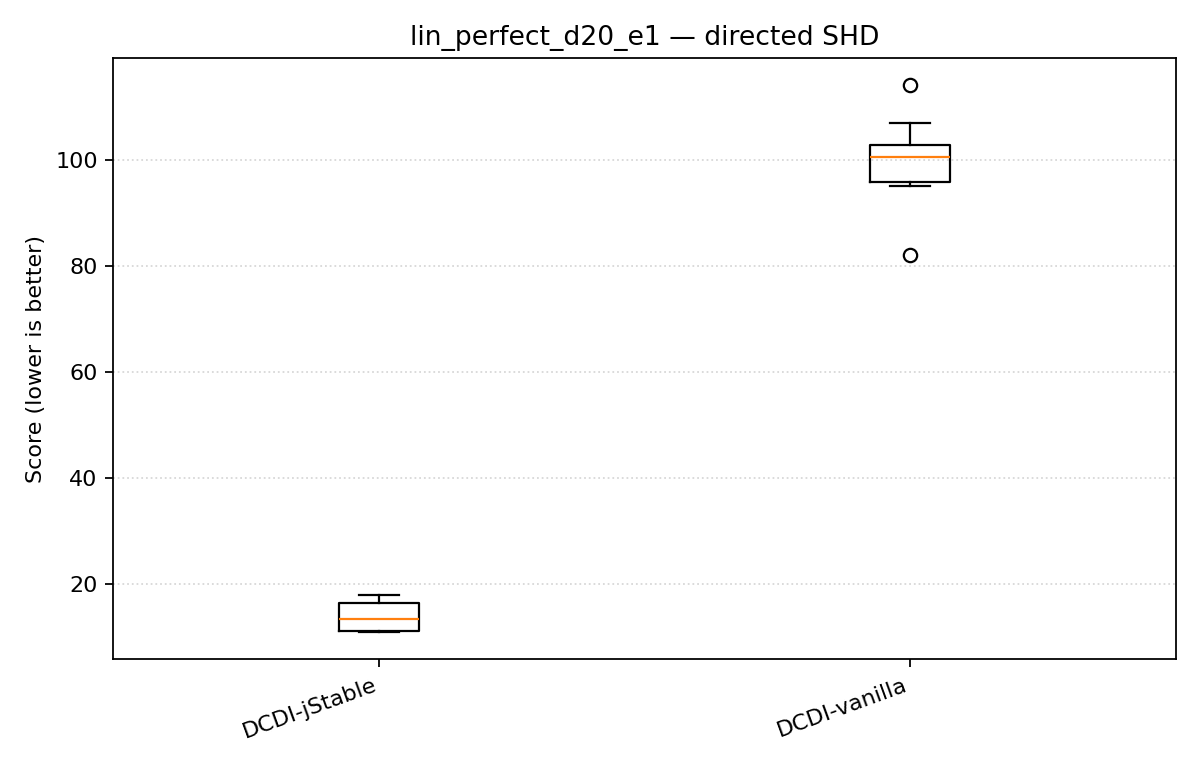}}
\hfill
\subfloat[\(d=20,e=1\), skeleton]{\includegraphics[width=.48\linewidth]{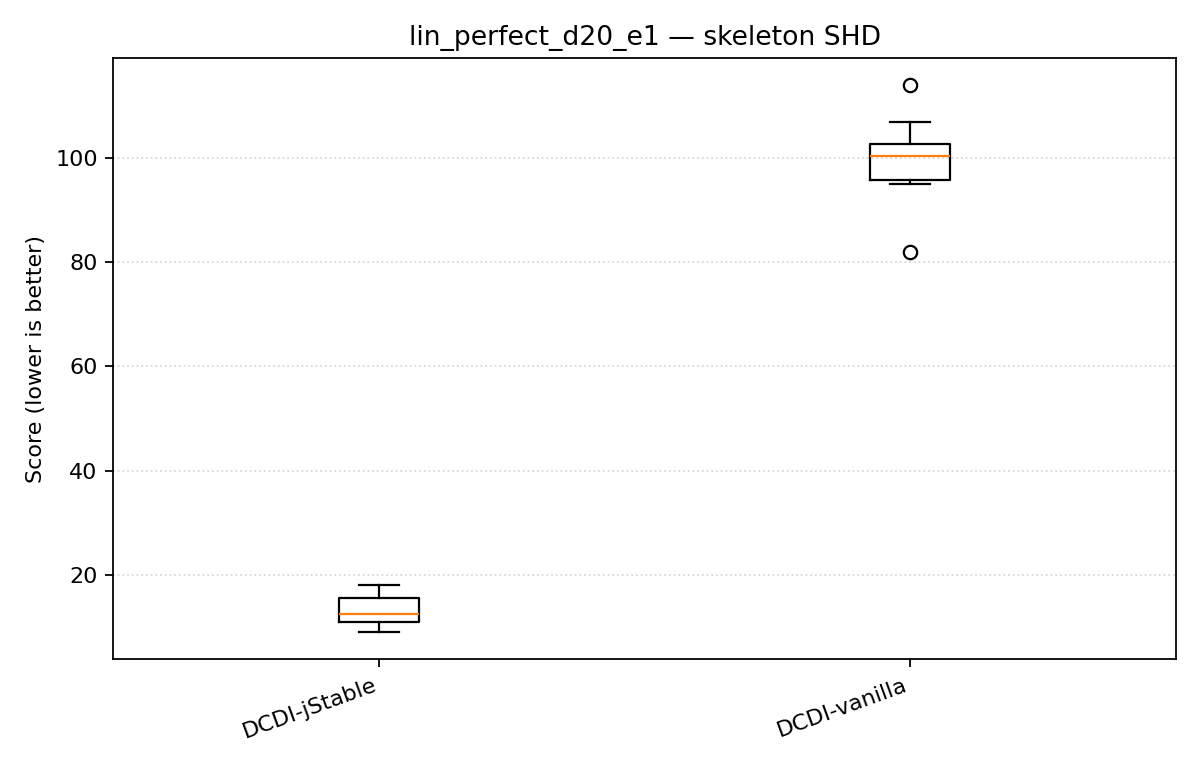}}\\
\subfloat[\(d=20,e=4\), directed]{\includegraphics[width=.48\linewidth]{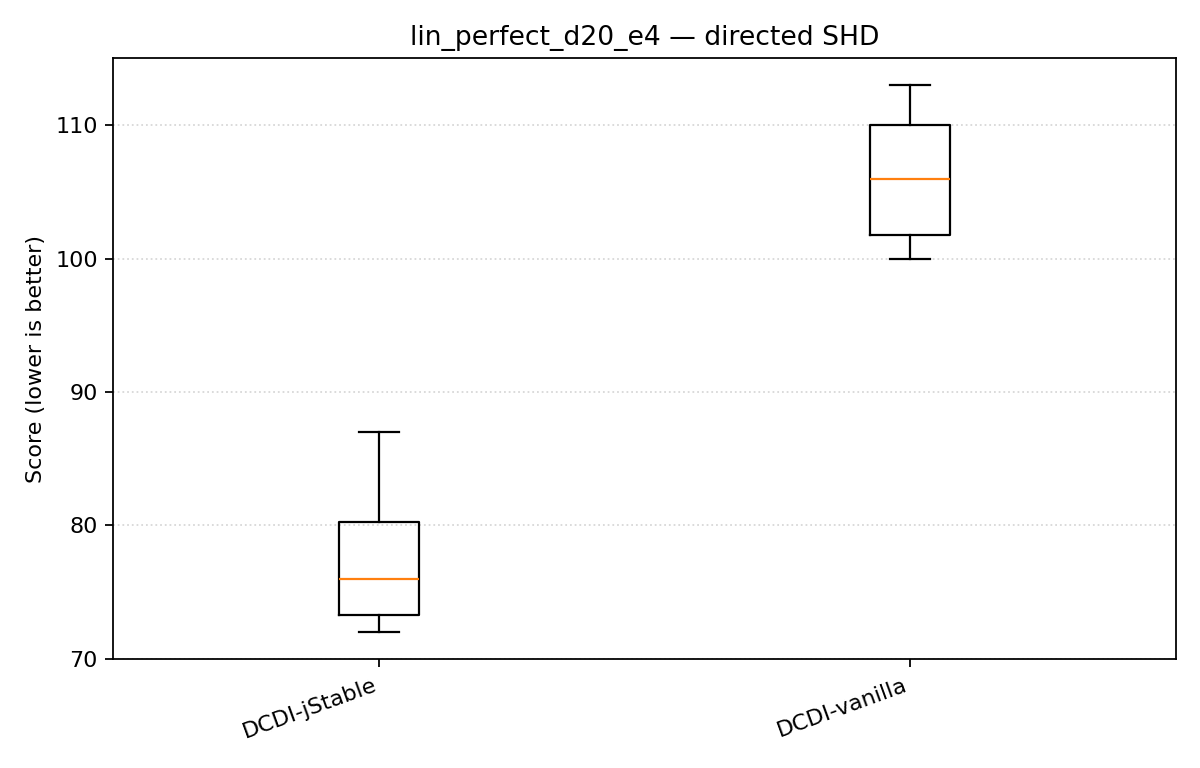}}
\hfill
\subfloat[\(d=20,e=4\), skeleton]{\includegraphics[width=.48\linewidth]{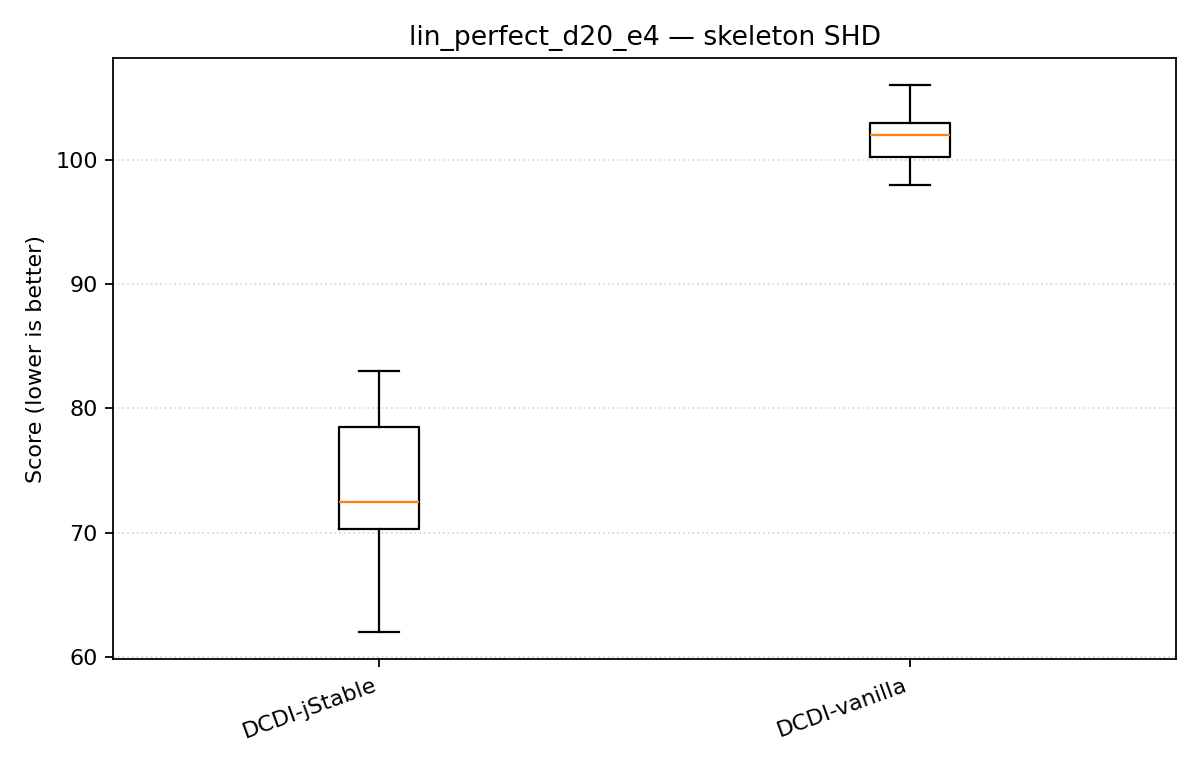}}
\caption{DCDI results for the \(d=20\) conditions, ten random graphs per
condition.}
\label{fig:dcdi20}
\end{figure}

\subsection{Sachs}

\begin{figure}[t]
\centering
\subfloat[GES]{\includegraphics[width=.32\linewidth]{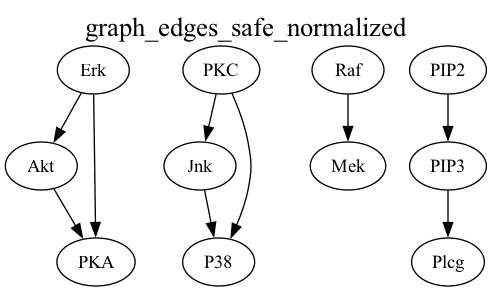}}
\subfloat[CGES]{\includegraphics[width=.32\linewidth]{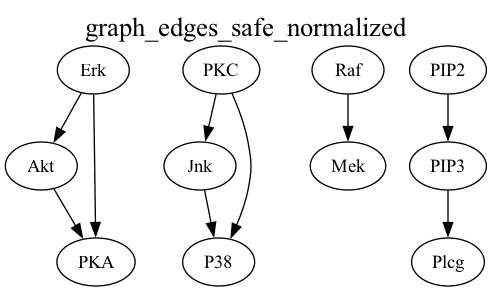}}
\subfloat[TCES]{\includegraphics[width=.32\linewidth]{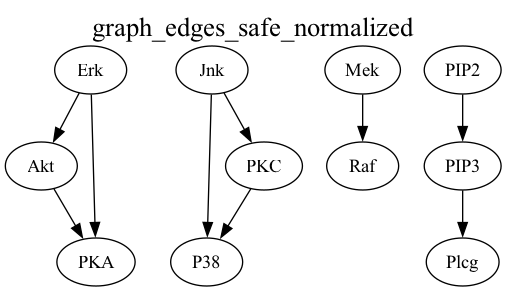}}
\caption{Learned graphs on the Sachs data.  The published Sachs network is a
reference rather than an oracle used for model selection.}
\label{fig:sachs}
\end{figure}

Figure~\ref{fig:sachs} compares GES, its structural regularization, and the
environment-regularized score.  These graphs illustrate how the penalties alter
the selected structure.

The \(\psi\)-FCI comparison in Table~\ref{tab:sachs-fci} is a useful negative
result.  The pooled estimator has higher \(F_1\) against the published reference,
while intersection reduces SHD for two thresholds but also removes supported
edges.  Stability aggregation is therefore not uniformly superior.

\begin{table}[t]
\centering
\caption{Sachs skeleton results: pooled versus intersection and all-but-1.
Higher \(F_1\) and lower SHD are preferred.}
\label{tab:sachs-fci}
\begin{tabular}{rcccccc}
\toprule
& \multicolumn{3}{c}{\(F_1\)}
& \multicolumn{3}{c}{SHD}\\
\(\alpha\) & Pooled & Inter. & All-but-1 & Pooled & Inter. & All-but-1\\
\midrule
0.01  & 0.514 & 0.370 & 0.370 & 17 & 17 & 17\\
0.02  & 0.486 & 0.370 & 0.370 & 19 & 17 & 17\\
0.005 & 0.485 & 0.320 & 0.320 & 17 & 17 & 17\\
\bottomrule
\end{tabular}
\end{table}

\subsection{Interference illustration}

\begin{figure}[t]
\centering
\subfloat[Edge frequencies by regime]{\includegraphics[width=.56\linewidth]{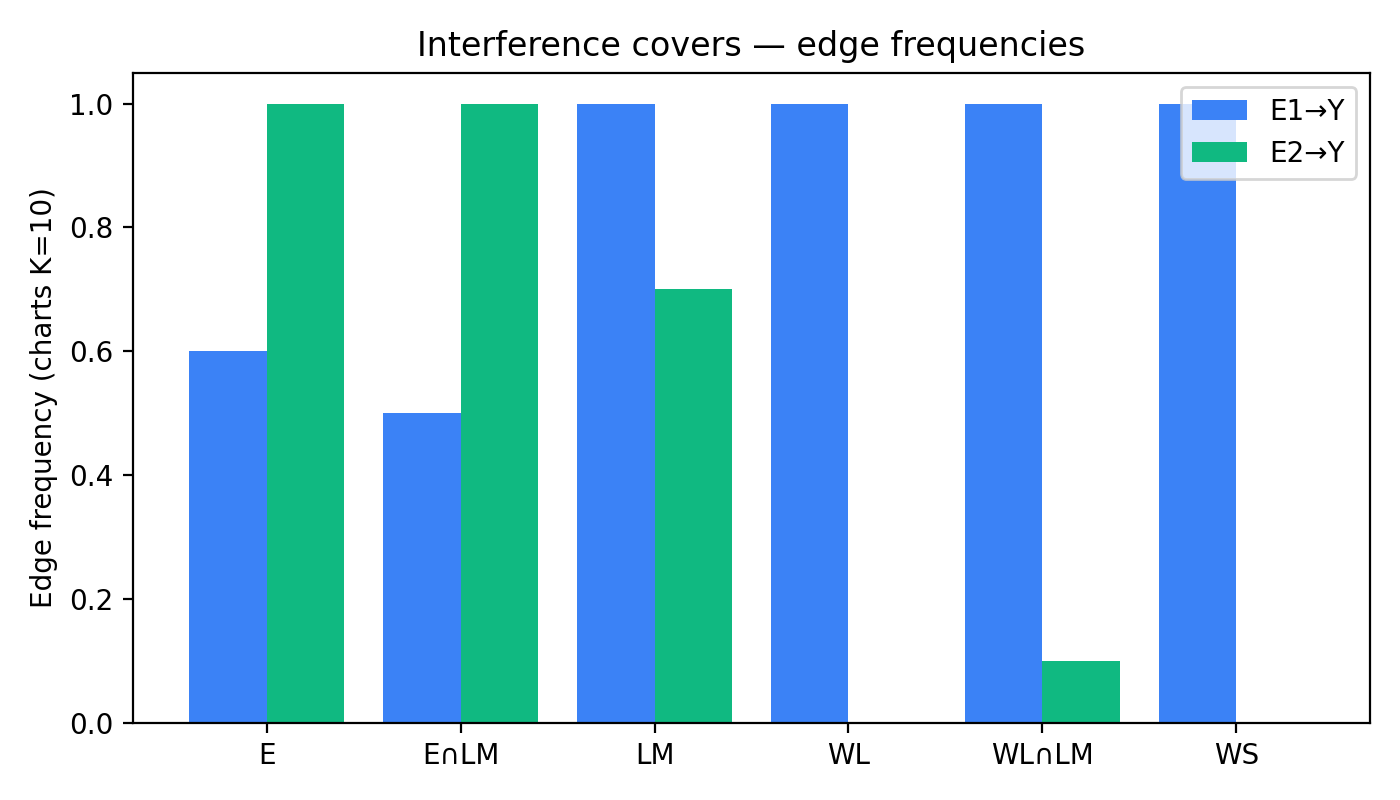}}
\hfill
\subfloat[Local coefficients]{\includegraphics[width=.40\linewidth]{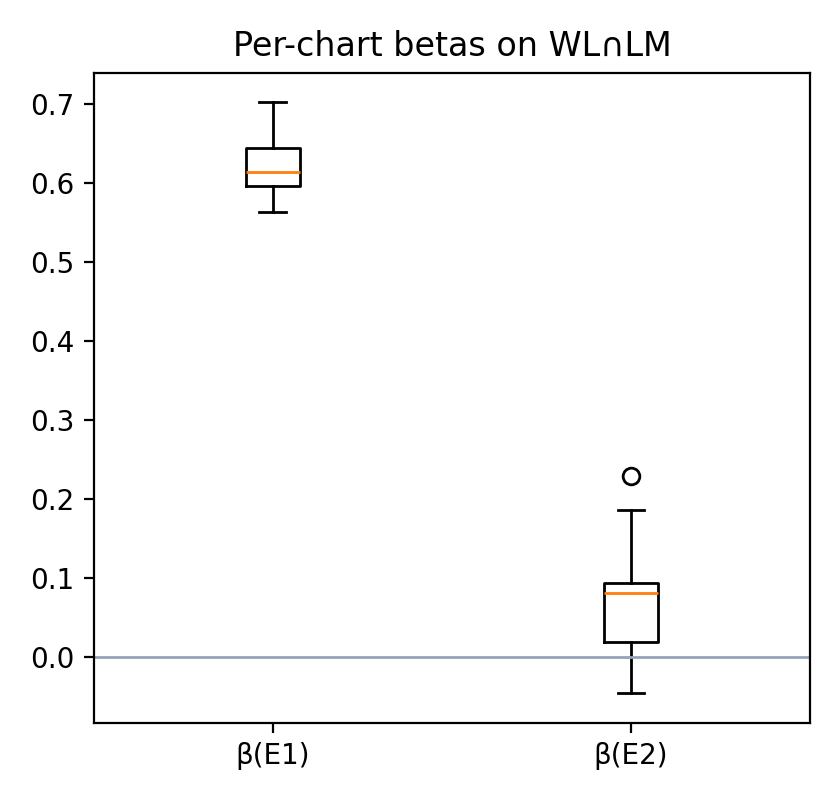}}
\caption{Two-source interference illustration with overlapping meteorological
regimes.}
\label{fig:interference}
\end{figure}

The two-source simulation partitions observations by wind direction and mixing
conditions.  Local regressions estimate the contributions \(E_1\to Y\) and
\(E_2\to Y\).  Figure~\ref{fig:interference} shows that the selected source
changes with the meteorological regime and that intersections retain the local
pattern.  This demonstrates context-dependent support; it is not itself an
identified policy effect under interference.  A causal policy analysis would
also require an exposure mapping and the assumptions appropriate to interference
\citep{wikle}.

\subsection{LINCS and PISA}

For the LINCS A375 subset, the union contains 428 directed candidates,
all-but-1 retains 130, and intersection retains 14.  The corresponding Jaccard
overlaps with the union are 0.304 and 0.033.  These numbers quantify how
aggressively the thresholds filter the graph.  Without a reference network they
do not establish that the retained edges are causal.

\begin{figure}[t]
\centering
\subfloat[Edge frequencies]{\includegraphics[width=.48\linewidth]{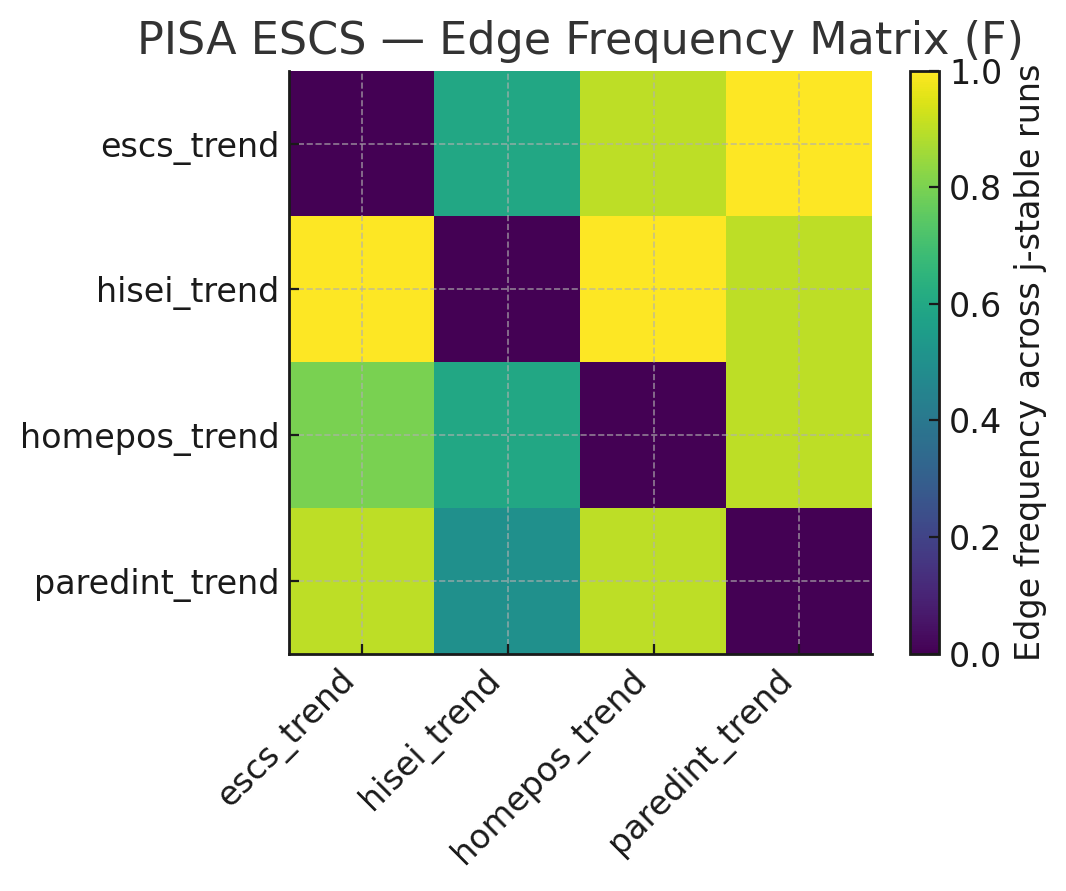}}
\hfill
\subfloat[Thresholded skeleton]{\includegraphics[width=.48\linewidth]{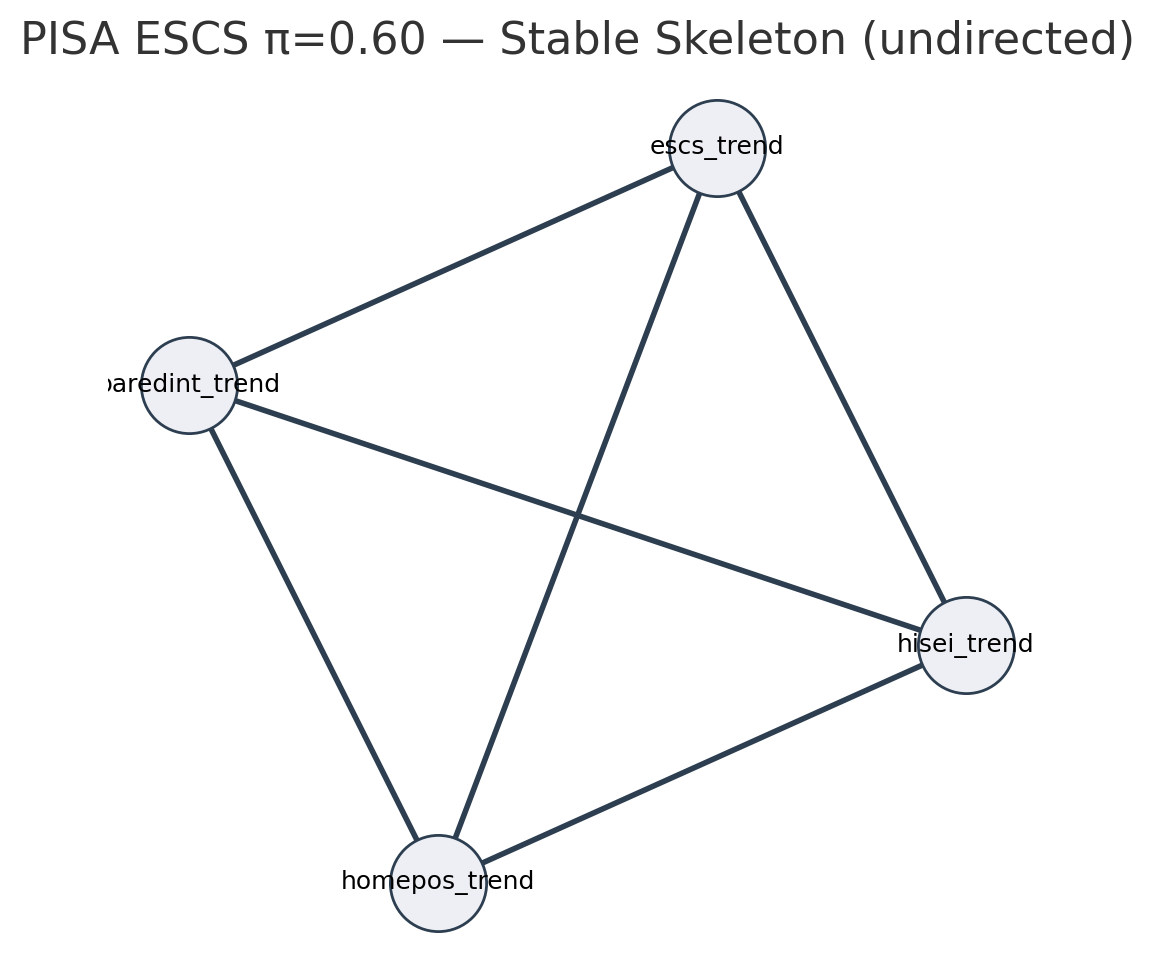}}
\caption{PISA ESCS pipeline sanity check.}
\label{fig:pisa}
\end{figure}

The PISA sanity check yields high-frequency relations between the three
components and their constructed ESCS composite.  Figure~\ref{fig:pisa} shows
the frequency matrix and thresholded skeleton.  Direction toward the composite
is partly supplied by the domain guard and should not be counted as discovered
orientation.

\clearpage

\section{Limitations and Practical Guidance}
\label{sec:limits}

\paragraph{A cover is a modeling choice.}
The result is conditional on which regimes are declared comparable.  Covers
should be specified from scientific knowledge or selected on training data and
then audited for overlap and sample size.

\paragraph{Stability is not causality.}
A stable confounded association can recur in every environment, while a true
edge can become hard to detect after an intervention or covariate shift.
Support aggregation inherits the assumptions and failure modes of the base
learner.

\paragraph{Direct interventions require masks.}
For a target-aware common-DAG analysis, use Eq.~\eqref{eq:masked} or a joint
interventional learner.  Unmasked intersection estimates persistence in the
mutilated regime graphs, which is a different target.

\paragraph{Do-calculus requires graph-surgery premises.}
An adjustment or action/observation identity should be derived chartwise using
the correct mutilated graph and then descended under the compatibility
hypotheses of \citet{sm:judo_calc}.  Neither an arbitrary probability aggregator
nor a high-frequency graph replaces that proof.

\paragraph{Compute comparisons need two budgets.}
Report both total CPU/GPU work and wall-clock time at a stated worker count.
Smaller local jobs may be faster and more memory efficient, but the collection
of jobs is not free.

\paragraph{Reproducibility.}
The arXiv source bundle contains the plotted artifacts but not a complete
executable pipeline or raw result archive.  Accordingly, the present paper
describes the archived measurements without claiming independent
reproducibility from this bundle alone.  A complete release should include data
provenance, environment manifests, target masks, per-regime adjacencies,
hyperparameter selection logs, and random seeds.

\section{Conclusion}

Decentralized causal discovery across regimes is useful even without a broad
categorical universality claim.  Local fitting exposes heterogeneity, support
aggregation provides an auditable stability filter, and the local computations
can be parallelized or performed behind data-silo boundaries.  The synthetic
experiments show cases in which this filter sharply improves SHD, while the
Sachs results demonstrate that it can also reduce useful signal.

The correct theoretical relationship to \(j\)-do-calculus is limited and clear:
compatible chartwise do-calculus derivations can descend to a covered context.
The empirical algorithms help propose and assess stable local structure, but do
not manufacture the causal premises.  This separation leaves a shorter and more
defensible contribution: a practical multi-environment discovery framework,
its intervention-aware correction, and an honest empirical account of when
stability aggregation helps.

\section{Changes in This Revision}
\label{sec:changes}

This version makes the following substantive changes:
\begin{enumerate}[leftmargin=*,itemsep=2pt]
  \item removes the proposed right-Kan causal-discovery estimator and associated
        universality, Markov-equivalence, and consistency claims;
  \item removes the claimed general Kan-extension interpretation of
        conditioning and intervention;
  \item delegates the categorical theory to the revised \(j\)-do-calculus paper
        and retains only its explicit local-to-global interface;
  \item replaces arbitrary aggregation of interventional probabilities with
        regime-wise reporting or a pre-specified convex mixture;
  \item distinguishes empirical support aggregation from literal sheafification
        and from proof of a do-calculus premise;
  \item introduces intervention-aware, child-masked edge support;
  \item replaces the former all-but-\(k\) consistency claim with a standard
        concentration statement under explicit selection assumptions;
  \item consolidates duplicated algorithmic and experimental sections;
  \item corrects and tempers the interpretation of the Sachs, LINCS, PISA,
        timing, and synthetic results; and
  \item records that the archived plots used the original unmasked aggregation,
        so they are not retrospectively presented as target-aware common-DAG
        estimates.
\end{enumerate}


\begin{thebibliography}{12}

\bibitem[Brouillard et al.(2020)]{dcdi}
Philippe Brouillard, S{\'e}bastien Lachapelle, Alexandre Lacoste, Simon
Lacoste-Julien, and Alexandre Drouin.
\newblock Differentiable causal discovery from interventional data.
\newblock In \emph{Advances in Neural Information Processing Systems}, 2020.

\bibitem[Chickering(2002)]{chickering:jmlr}
David Maxwell Chickering.
\newblock Optimal structure identification with greedy search.
\newblock \emph{Journal of Machine Learning Research}, 3:507--554, 2002.

\bibitem[Evangelista et al.(2022)]{lincs1000}
John~E. Evangelista et al.
\newblock SigCom LINCS: data and metadata search engine for a million gene
expression signatures.
\newblock \emph{Nucleic Acids Research}, 50(W1):W697--W709, 2022.

\bibitem[Jaber et al.(2020)]{psifci}
Amin Jaber, Jiji Zhang, and Elias Bareinboim.
\newblock Causal identification under Markov equivalence: completeness results.
\newblock In \emph{Proceedings of ICML}, 2020.

\bibitem[Mac Lane and Moerdijk(1992)]{maclane1992sheaves}
Saunders Mac Lane and Ieke Moerdijk.
\newblock \emph{Sheaves in Geometry and Logic}.
\newblock Springer, 1992.

\bibitem[Mahadevan(2025)]{sm:judo_calc}
Sridhar Mahadevan.
\newblock Intuitionistic \(j\)-do-calculus in topos causal models.
\newblock arXiv:2510.17944, revised version, 2025.

\bibitem[Parascandolo et al.(2018)]{icm}
Giambattista Parascandolo, Niki Kilbertus, Mateo Rojas-Carulla, and Bernhard
Sch{\"o}lkopf.
\newblock Learning independent causal mechanisms.
\newblock In \emph{Proceedings of ICML}, 2018.

\bibitem[Pearl(2009)]{pearl-book}
Judea Pearl.
\newblock \emph{Causality: Models, Reasoning, and Inference}.
\newblock Cambridge University Press, second edition, 2009.

\bibitem[Sachs et al.(2005)]{sachs}
Karen Sachs, Omar Perez, Dana Pe'er, Douglas~A. Lauffenburger, and Garry~P.
Nolan.
\newblock Causal protein-signaling networks derived from multiparameter
single-cell data.
\newblock \emph{Science}, 308(5721):523--529, 2005.

\bibitem[Wikle and Zigler(2023)]{wikle}
Nathan~B. Wikle and Corwin~M. Zigler.
\newblock Causal health impacts of power plant emission controls under modeled
and uncertain physical process interference.
\newblock arXiv:2306.05665, 2023.

\end{thebibliography}
\end{document}